\documentclass[letterpaper, 10 pt, conference]{ieeeconf}  

\IEEEoverridecommandlockouts                              
\usepackage{epsfig} 
\usepackage{booktabs}
\usepackage{graphicx}
\usepackage[style=ieee,sorting=none,doi=false,url=true,minbibnames=1,maxbibnames=1]{biblatex}
\title{\LARGE \bf
    Skeleton-Based Posture Classification to Promote Safer Walker-Assisted Gait in Older Adults
}

\author{Sergio D. Sierra M.$^{1,2,*}$, Monica Sinha$^{1,2,*}$, Marcela Múnera$^{1}$, Carlos A. Cifuentes$^{1}$
\thanks{Correspondence Author: sergio.sierramarin@uwe.ac.uk}
\thanks{Partially funded by the EPSRC Farscope CDT and MRC MR/Y010620/1.}
\thanks{$^{1}$ Sergio S., Mónica M., Marcela M., Carlos C., are with the Bristol Robotics Laboratory, University of the West of England, Bristol, UK.}%
\thanks{$^{2}$ Faculty of Engineering, University of Bristol, Bristol, UK.}%
\thanks{* These authors equally contributed to this work.}
}

\begin{document}

\maketitle
\thispagestyle{empty}
\pagestyle{empty}

\begin{abstract}

Falls among older adults are a significant public health concern, leading to severe injuries, loss of independence, and increased healthcare costs. This study evaluates the effectiveness of various models, including a Geometric approach, XGBoost, SVM, and several deep learning architectures, in classifying walker usage, standing vs. sitting, and posture for smart walkers used. Geometric and XGBoost were the top performers. XGBoost achieved near-perfect training accuracy in binary classification tasks, with 99.84\% for walker choice and 99.69\% for standing vs. sitting. For posture classification, Geometric approach attained 89.9\% accuracy for 8 postures, and XGBoost obtained 99.24\% during training for 17 postures. Deep learning models such as the 4-layer CNN and Encoder-Decoder CNN also demonstrated strong performance in binary classification, with accuracies above 98\%. This study underscores the potential of machine learning to enhance human-robot interaction in smart walkers, particularly for fall prevention.

\end{abstract}

\section{Introduction}
The global population is aging at an unprecedented rate, leading to an increase in the number of older adults worldwide. Projections indicate that the number of individuals aged 65 and older is expected to rise from 10\% in 2022 to 16\% in 2050 \cite{UnitedNations2019}. This demographic shift brings about various health challenges, including declines in vision, balance, loss of muscle mass, mobility, and cognitive functions \cite{Martins2024}. One of the most critical issues arising from these complications is the increased risk of falls, a leading cause of injuries, loss of independence, and mortality among older adults \cite{NIA2022}. 

To mitigate these risks, assistive devices such as walkers and canes have traditionally been employed to provide support and enhance balance. These devices are affordable and offer a simple mechanical structure capable of providing weight support, increased base of support, improved lateral stability, and safer walking \cite{Sakano2023}. However, the misuse of such devices can paradoxically increase the risk of falls. Incorrect posture or improper use of a walker can lead to a loss of balance, resulting in a fall \cite{Gell2015, Resnik2009}. Therefore, it is essential to  correct the posture of older adults using assistive devices.

Advancements in robotics and marker-less motion tracking technology offer a promising solution to these challenges~\cite{Stefana2021, Chung2022}. Traditional motion capture systems require physical markers attached to the body, which can be intrusive for older adults. Marker-less motion tracking eliminates this need, allowing for the seamless monitoring of movement without physical interference. These technologies include inertial measurement units (IMUs), smart insoles, smart watches, and camera-based systems \cite{Huang2023}. Despite their effectiveness, the adoption of these technologies among older adults remains limited. High costs, complex usability, and the tendency of older adults to forget or improperly use wearable devices are significant barriers to widespread use.

Current approaches for posture classification primarily focus on general activities classification \cite{Ogundokun2022}. These solutions are often based on datasets with full-body views of users and require substantial processing capacities, which may not be feasible for real-time applications in assistive devices \cite{Wang2021, Rababaah2022}. Given these challenges, posture classification using on-board cameras mounted on assistive devices like four-wheeled walkers presents a promising alternative. Such systems can continuously monitor the user's posture without requiring the individual to wear additional devices. 

\begin{figure}[t]
    \centering
    \includegraphics[scale=0.9]{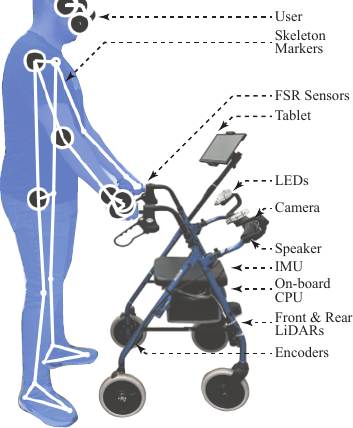}
    \caption{Smart Walker with sensors and feedback systems.}
    \label{fig:walker}
    \vspace{-0.4cm}
\end{figure}

\begin{figure*}
    \vspace{0.2cm}
    \centering
    \includegraphics[scale=1.1]{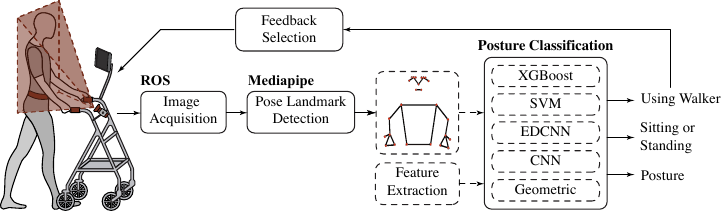}
    \caption{Outline of the image acquisition process for pose landmark detection and further posture classification. eXtreme Gradient Boosting (XGBoost), Support Vector Machine (SVM), Encoder Decoder Convolutional Neural Network (EDCNN), Convolutional Neural Network (CNN).}
    \label{fig:diagram}
    \vspace{-0.3cm}
\end{figure*}

This article compares multiple techniques for posture classification using both supervised learning and traditional techniques. A dataset of full-body and upper-body postures was captured from walker-mounted and external cameras (See Fig. \ref{fig:walker}). The training and implementation of the algorithms for posture classification are detailed. This method could reduce the risk of falls among older adults by detecting improper postures and alerting in advance. The potential applications in settings like care homes and private residences are discussed. This work addresses limitations of existing methods, enhancing the privacy, safety, and mobility of older adults and ultimately improving their quality of life.

\section{Methodology}
The methodology is shown in Fig.~\ref{fig:diagram}. In this work, five models are proposed and compared. The diagram illustrates the process for image acquisition, landmark detection, feature extraction, posture classification, and feedback.

\subsection{Smart Device and Acquisition System}
The Socially Assistive Walker (SAW) \cite{Sierra2024} (Fig. \ref{fig:walker}) is a conventional walker frame enhanced with sensors and feedback components. It runs on a Raspberry Pi 4 (8GB) with ROS and Mosquitto. The system includes a USB camera (C270, Logitech) for upper-body imaging, a USB speaker (HONKYOB) for auditory feedback, LED NeoPixel rings (Adafruit) and a tablet (ELECROW) for visual feedback. Force Sensing Resistors (FSRs, Interlink Electronics) monitor grip pressure. A front LiDAR (RPLiDAR A2, SLAMTEC) detects obstacles, while a rear LiDAR (URG-04LX-UG01, Hokuyo) tracks leg movement. An Inertial Measurement Unit (BNO080, Sparkfun) and magnetic encoders (AS5600, Osram) monitor odometry. A tripod-mounted USB camera was used for full-body acquisition.

\subsection{Pose Landmark Detection}
Google MediaPipe's Pose Detection framework was used to track key body landmarks in real-time video streams. It employs a two-step process: a pose detection model identifies human bodies, followed by a pose landmarker model that maps 33 three-dimensional landmarks~\cite{google}. Built on a MobileNetV2-based CNN, this BlazePose variant utilizes GHUM for full 3D body pose estimation~\cite{google}.  

This study uses pose landmarks as point coordinates instead of raw images for training posture classification models, reducing the need for large, diverse datasets. MediaPipe’s Pose Landmarker (lite) enables local execution on Raspberry Pi 4, with cloud-based processing available if needed, transmitting only text-based landmarks for efficiency and security.

\subsection{Dataset Collection}
Data was collected for training purposes of the supervised learning models. This was conducted under controlled conditions at the Bristol Robotics Laboratory (BRL), using the previously described system. Participants performed a series of predefined exercises such as arm raises, elbow folds, torso bends, neck movements, as well as standing and sitting postures. Breaks were offered between exercises to prevent fatigue. The inclusion criteria were healthy individuals, between the ages of 18 and 59, who could understand instructions, were willing to voluntarily provide informed consent, and had the ability to perform the required exercises. The exclusion criteria were individuals with severe cardiovascular issues, mobility impairments, or cognitive impairments.

Data was collected from 21 participants (age range: 21–48 years), of which 17 participants' data were used for training (80\%) and testing (20\%) the model. The remaining 4 participants were used to predict the model's performance in a real-time application with unseen data. To ensure accurate data labeling, a graphical user interface (GUI) was used. The GUI was developed with the Tkinter library in Python 3.8. The GUI allowed input of participants' ID, selection of upper- or full-body view, posture type, preview video capture, and initiation of data recording. For the purpose of this study, only markers' data were stored (i.e., text data).

\subsection{Feature Extraction}
Prior to model training, data was pre-processed to transform raw data (e.g., pose landmark x, y, z coordinates) into more informative features. In this study, 48 features such as distances and angles between joints were extracted. 

\subsection{Posture Classification}
Four types of models were selected and trained with full and upper body marker data, and one additional approach was implemented without training requirements. The supervised learning models predict three distinct outputs: Choice of Walker, a binary classification output determining if a walker is used; Initial Position, a binary classification output identifying whether the posture begins from a sitting or standing position; and Posture Classification, a multi-class classification output identifying 17 postures.

\subsubsection{Convolutional Neural Network (CNN)} In this work, CNNs were used by means of a 4-layer and a 6-layer CNN. For the 4-layer CNN, two variations of input layers were tested with feature-engineered data (i.e., 48x1) and with raw pose landmarks data (i.e., 33x3). For the 6-layer CNN, only raw data was used as input. The CNN layers apply filters to the input, followed by Batch Normalization to stabilize learning and Max Pooling to reduce spatial dimensions. The pooling layers down-sample the data, retaining important features while reducing dimensionality, which helps in efficiently extracting movement patterns. After convolution and pooling, the data is flattened into a 1D vector. This vector passes through several fully connected (dense) layers, followed by dropout layers to prevent overfitting \cite{OgundokunCNN}.

\subsubsection{Encoder Decoder CNN (EDCNN)} The Encoder-Decoder architecture is useful for tasks that require both dimensionality reduction and output reconstruction, such as image segmentation, sequence-to-sequence tasks, or multi-output predictions. The encoder compresses the input data through a series of convolutional and max-pooling layers, which extract features and reduce spatial dimensions while retaining important information. In this model, the encoder includes 3 convolutional layers with 64, 128, and 256 filters, each of them followed by max-pooling layers. The decoder reconstructs the spatial feature maps using transposed convolutional and up-sampling layers \cite{Yasrab2017}. After up-sampling, the feature maps are flattened and passed to fully connected layers for classification. The final architecture follows an encoder-decoder-classifier sequence, where the encoder extracts features, the decoder refines the representation, and the dense layers classify the outputs. Unlike conventional autoencoder-style training, the encoder and decoder are trained jointly for classification rather than being pre-trained separately. 

\subsubsection{Support Vector Machine (SVM)} SVMs are supervised machine-learning algorithms utilized for both classification and regression tasks. The core principle behind SVM is to identify the optimal hyperplane that maximizes the margin between distinct classes \cite{bdcc8020013}. To predict the previously described outputs, a separate SVM model is used. 

\subsubsection{Extreme Gradient Boosting (XGBoost)} XGBoost is a machine learning algorithm that is able to handle complex data patterns through gradient boosting. It builds an ensemble of decision trees, with each tree aiming to correct the errors of its predecessors, thereby creating a model capable of capturing intricate relationships within the data~\cite{bentejac2021comparative}. The XGBoost classifier was combined with a \textit{MultiOutputClassifier}, enabling it to predict several variables simultaneously from the extracted features and raw pose landmarks.

During preliminary evaluations, XGBoost demonstrated superior performance compared to the other models. Given the goal to identify potentially hazardous postures and risk factors, the XGBoost model was also trained on both full-body and upper-body data and with a simplified output to a single classification task with three possible labels: "\textit{standing}", "\textit{sitting}", or "\textit{bad posture}". 

\subsection{Geometric Approach}
The last model focuses on a simpler approach that just analyzes the posture landmarks' geometric features. A baseline is established with an initial pose calibration. To classify the posture, deviations of any given pose from the initial calibrated pose were computed using a vectorized approach. The Euclidean distances between corresponding landmarks of the initial and given poses were calculated. This process yielded a deviation vector for each landmark. 

Afterwards, a weighted posture classification model was developed based on the calculated deviation vectors. The model evaluated the deviations against empirically determined thresholds to classify various postures such as standing, leaning forward, leaning backward, twisting left, twisting right, lifting hands, and sitting (i.e., 8 poses). Additionally, the embedded force sensors on the walker handles were used to contribute to lifting hand postures. The thresholds were established through preliminary experiments and observations. The classification criteria involved assessing the deviations of specific key landmarks, such as the nose, shoulders, and hands, to determine the subject's posture.

\section{Results}
Data was recorded and labeled using the SAW Raspberry Pi at a rate of 10 FPS. Model training was done using an external computer AORUS 15 XE4, Gigabyte with 12th Gen Intel Core i7-12700H (20 CPUs), 16Gb of RAM, NVIDIA GeForce RTX 3070 Ti 16Gb, and Intel Iris Xe Graphics 8Gb.

\subsection{Training and Testing Metrics for Models}

\begin{table}[t]
\vspace{0.2cm}
\caption{Metrics for 4-layer CNN with raw pose landmark data}
\label{tab:4layercnn-raw}
\resizebox{\columnwidth}{!}{%
\begin{tabular}{@{}lccccc@{}}
\toprule
 & \multicolumn{5}{c}{\textbf{Training and Validation}} \\ \midrule
\textbf{Output Label} & \multicolumn{1}{l}{\textbf{Accuracy}} & \multicolumn{1}{l}{\textbf{Precision}} & \multicolumn{1}{l}{\textbf{F1 Score}} & \multicolumn{1}{l}{\textbf{Sensitivity}} & \multicolumn{1}{l}{\textbf{Specificity}} \\
Choice of Walker & 0.997 & 0.993 & 0.995 & 0.997 & 0.688 \\
Initial Position & 0.989 & 0.993 & 0.993 & 0.993 & 0.219 \\
Posture Type & 0.900 & 0.925 & 0.902 & 0.880 & 0.995 \\ \midrule
\textbf{Output Label} & \multicolumn{5}{c}{\textbf{Prediction}} \\ \midrule
Choice of Walker & 0.967 & 1.000 & 0.955 & 0.914 & 1.000 \\
Initial Position & 0.963 & 0.973 & 0.980 & 0.987 & 0.606 \\
Posture Type & 0.760 & 0.850 & 0.764 & 0.756 & 0.984 \\ \bottomrule
\end{tabular}%
}
\end{table}

The results for the 4-layer Convolutional Neural Network (CNN) using raw pose landmark data are shown in Table \ref{tab:4layercnn-raw}. This model exhibited high accuracy and precision across the different output labels during training and validation. Specifically, for the choice of walker, the accuracy reached 99.7\%, while the initial position and posture type also demonstrated strong performance with accuracies of 98.9\% and 90.0\%, respectively. However, the sensitivity and specificity varied significantly, particularly for the initial position, which had lower specificity (0.219) compared to other metrics.

\begin{table}[t]
\vspace{0.2cm}
\caption{Training and prediction metrics for 4-layer CNN with raw pose landmark data and extracted features}
\label{tab:4layercnn-feat}
\resizebox{\columnwidth}{!}{%
\begin{tabular}{@{}lccccc@{}}
\toprule
 & \multicolumn{5}{c}{\textbf{Training and Validation}} \\ \midrule
\textbf{Output Label} & \multicolumn{1}{l}{\textbf{Accuracy}} & \multicolumn{1}{l}{\textbf{Precision}} & \multicolumn{1}{l}{\textbf{F1 Score}} & \multicolumn{1}{l}{\textbf{Sensitivity}} & \multicolumn{1}{l}{\textbf{Specificity}} \\
Choice of Walker & 0.988 & 0.979 & 0.992 & 0.982 & 0.688 \\
Initial Position & 0.971 & 0.976 & 0.993 & 0.988 & 0.212 \\
Posture Type & 0.810 & 0.880 & 0.900 & 0.752 & 0.992 \\ \midrule
\textbf{Output Label} & \multicolumn{5}{c}{\textbf{Prediction}} \\ \midrule
Choice of Walker & 0.998 & 1.000 & 0.997 & 0.995 & 1.000 \\
Initial Position & 0.973 & 0.998 & 0.985 & 0.972 & 0.975 \\
Posture Type & 0.682 & 0.743 & 0.664 & 0.682 & 0.980 \\ \bottomrule
\end{tabular}%
}
\end{table}

When incorporating both raw pose landmark data and extracted features, the performance metrics slightly shifted, as detailed in Table \ref{tab:4layercnn-feat}. The 4-layer CNN maintained high accuracy for the choice of walker (98.8\%), for the initial position (97.1\%) and posture type (81.0\%), though there was a noticeable drop in accuracy for posture type compared to the raw data alone. The inclusion of extracted features helped the model to improve prediction accuracy for the choice of walker and initial position, achieving 99.8\% and 97.3\%.

\begin{table}[t]
\caption{Training and prediction metrics for 6-layer CNN with raw pose landmark data.}
\label{tab:6layercnn}
\resizebox{\columnwidth}{!}{%
\begin{tabular}{@{}lccccc@{}}
\toprule
 & \multicolumn{5}{c}{\textbf{Training and Validation}} \\ \midrule
\textbf{Output Label} & \multicolumn{1}{l}{\textbf{Accuracy}} & \multicolumn{1}{l}{\textbf{Precision}} & \multicolumn{1}{l}{\textbf{F1 Score}} & \multicolumn{1}{l}{\textbf{Sensitivity}} & \multicolumn{1}{l}{\textbf{Specificity}} \\
Choice of Walker & 0.990 & 0.984 & 0.985 & 0.986 & 0.689 \\
Initial Position & 0.978 & 0.988 & 0.986 & 0.984 & 0.221 \\
Posture Type & 0.853 & 0.870 & 0.860 & 0.850 & 0.996 \\ \midrule
\textbf{Output Label} & \multicolumn{5}{c}{\textbf{Prediction}} \\ \midrule
Choice of Walker & 0.993 & 0.998 & 0.991 & 0.983 & 0.999 \\
Initial Position & 0.931 & 0.998 & 0.962 & 0.929 & 0.975 \\
Posture Type & 0.757 & 0.806 & 0.747 & 0.757 & 0.984 \\ \bottomrule
\end{tabular}%
}
\end{table}

The 6-layer CNN results, presented in Table \ref{tab:6layercnn}, showed overall strong performance, with particularly high specificity for posture type (99.6\%). However, its accuracy and precision for the posture type were relatively lower compared to other metrics, indicating some challenges in classification.

\begin{table}[t]
\vspace{0.2cm}
\caption{Training and prediction metrics for EDCNN with raw pose landmark data.}
\label{tab:edcnn}
\resizebox{\columnwidth}{!}{%
\begin{tabular}{@{}lccccc@{}}
\toprule
 & \multicolumn{5}{c}{\textbf{Training and Validation}} \\ \midrule
\textbf{Output Label} & \multicolumn{1}{l}{\textbf{Accuracy}} & \multicolumn{1}{l}{\textbf{Precision}} & \multicolumn{1}{l}{\textbf{F1 Score}} & \multicolumn{1}{l}{\textbf{Sensitivity}} & \multicolumn{1}{l}{\textbf{Specificity}} \\
Choice of Walker & 0.996 & 0.992 & 0.993 & 0.995 & 0.684 \\
Initial Position & 0.992 & 0.996 & 0.995 & 0.994 & 0.214 \\
Posture Type & 0.908 & 0.928 & 0.911 & 0.895 & 0.995 \\ \midrule
\textbf{Output Label} & \multicolumn{5}{c}{\textbf{Prediction}} \\ \midrule
Choice of Walker & 0.971 & 1.000 & 0.961 & 0.925 & 1.000 \\
Initial Position & 0.930 & 0.981 & 0.962 & 0.944 & 0.729 \\
Posture Type & 0.660 & 0.648 & 0.640 & 0.660 & 0.978 \\ \bottomrule
\end{tabular}%
}
\end{table}

The Encoder-Decoder CNN (EDCNN) results, detailed in Table \ref{tab:edcnn}, highlighted good accuracy and precision in training, with accuracy values of 99.6\% for the choice of walker, 99.2\% for initial position, and 90.8\% for posture type. The prediction metrics showed a drop in performance, especially for posture type, which had lower precision and F1 scores compared to training, 66.0\% and 64.0\%, respectively.

\begin{table}[t]
\caption{Training and prediction metrics for SVM with raw pose landmark data.}
\label{tab:svm}
\resizebox{\columnwidth}{!}{%
\begin{tabular}{@{}lccccc@{}}
\toprule
 & \multicolumn{5}{c}{\textbf{Training and Validation}} \\ \midrule
\textbf{Output Label} & \multicolumn{1}{l}{\textbf{Accuracy}} & \multicolumn{1}{l}{\textbf{Precision}} & \multicolumn{1}{l}{\textbf{F1 Score}} & \multicolumn{1}{l}{\textbf{Sensitivity}} & \multicolumn{1}{l}{\textbf{Specificity}} \\
Choice of Walker & 0.974 & 0.937 & 0.960 & 0.985 & 0.970 \\
Initial Position & 0.903 & 0.904 & 0.940 & 0.979 & 0.630 \\
Posture Type & 0.790 & 0.810 & 0.790 & 0.790 & 0.986 \\ \midrule
\textbf{Output Label} & \multicolumn{5}{c}{\textbf{Prediction}} \\ \midrule
Choice of Walker & 0.934 & 1.000 & 0.903 & 0.824 & 0.912 \\
Initial Position & 0.968 & 0.967 & 0.983 & 1.000 & 0.750 \\
Posture Type & 0.667 & 0.682 & 0.653 & 0.667 & 0.978 \\ \bottomrule
\end{tabular}%
}
\end{table}

The Support Vector Machine (SVM) model showed variable performance across different tasks. For training the choice of walker, it achieved an accuracy of 97.4\% with 93.7\% precision but had a higher sensitivity of 98.5\%. For initial position, the model reached a 90.3\% accuracy and precision of 90.4\%. However, for posture type classification, SVM's accuracy was lower at 79.0\%, with 81.0\% precision and 79.0\% sensitivity. For detailed metrics, see Table \ref{tab:svm}.

\begin{table}[t]
\vspace{0.1cm}
\caption{Training and prediction metrics for XGBoost}
\label{tab:xgboost}
\resizebox{\columnwidth}{!}{%
\begin{tabular}{@{}lccccc@{}}
\toprule
 & \multicolumn{5}{c}{\textbf{Training and Validation}} \\ \midrule
\textbf{Output Label} & \multicolumn{1}{l}{\textbf{Accuracy}} & \multicolumn{1}{l}{\textbf{Precision}} & \multicolumn{1}{l}{\textbf{F1 Score}} & \multicolumn{1}{l}{\textbf{Sensitivity}} & \multicolumn{1}{l}{\textbf{Specificity}} \\
Choice of Walker & 1.000 & 1.000 & 1.000 & 1.000 & 1.000 \\
Initial Position & 0.999 & 0.999 & 0.999 & 0.999 & 0.998 \\
Posture Type & 0.992 & 0.992 & 0.9992 & 0.992 & 1.000 \\ \midrule
\textbf{Output Label} & \multicolumn{5}{c}{\textbf{Prediction}} \\ \midrule
Choice of Walker & 0.998 & 1.000 & 0.997 & 0.995 & 1.000 \\
Initial Position & 0.996 & 0.998 & 0.998 & 0.998 & 0.975 \\
Posture Type & 0.741 & 0.749 & 0.719 & 0.740 & 1.000 \\ \bottomrule
\end{tabular}%
}
\end{table}

\begin{table}[t]
\caption{Training and prediction metrics for XGBoost for single label with multiple classification.}
\label{tab:xgboost2}
\resizebox{\columnwidth}{!}{%
\begin{tabular}{@{}lccccc@{}}
\toprule
 & \multicolumn{5}{c}{\textbf{Training and Validation}} \\ \midrule
\textbf{Output Label} & \multicolumn{1}{l}{\textbf{Accuracy}} & \multicolumn{1}{l}{\textbf{Precision}} & \multicolumn{1}{l}{\textbf{F1 Score}} & \multicolumn{1}{l}{\textbf{Sensitivity}} & \multicolumn{1}{l}{\textbf{Specificity}} \\
Single Output & 1.000 & 1.000 & 0.999 & 0.999 & 0.998 \\ \midrule
\textbf{Output Label} & \multicolumn{5}{c}{\textbf{Prediction}} \\ \midrule
Single Output & 0.850 & 0.830 & 0.830 & 0.850 & 0.829 \\ \bottomrule
\end{tabular}%
}

\end{table}

XGBoost's performance, as shown in Tables \ref{tab:xgboost} and \ref{tab:xgboost2}, was better than the other models. The model achieved 100\% accuracy and precision during training for all outputs. For prediction, XGBoost maintained high accuracy and precision for the choice of walker and initial position, although the performance for posture type showed a decline in accuracy and F1 score, reflecting some challenges in classifying posture types accurately. These results underscore the effectiveness of XGBoost in handling multiple output classifications and its superior performance in comparison to CNN models and SVM in this context. The XGBoost ability to handle both raw and processed data effectively highlights its robustness for posture classification tasks. When training XGBoost for a single label with multiple classifications, the training accuracy was 100\% and prediction accuracy was 85.0\%.

\begin{table}[t]
\caption{Prediction metrics for Geometric Approach.}
\label{tab:geometric}
\resizebox{\columnwidth}{!}{%
\begin{tabular}{@{}lccccc@{}}
\toprule
 & \multicolumn{5}{c}{\textbf{Prediction}} \\ \midrule
\textbf{Output Label} & \multicolumn{1}{l}{\textbf{Accuracy}} & \multicolumn{1}{l}{\textbf{Precision}} & \multicolumn{1}{l}{\textbf{F1 Score}} & \multicolumn{1}{l}{\textbf{Sensitivity}} & \multicolumn{1}{l}{\textbf{Specificity}} \\
Single Output & 0.899 & 0.9126 & 0.908 & 0.912 & 0.898 \\ \bottomrule
\end{tabular}%
}
\end{table}

Finally, three healthy users were asked to reproduce the 8 poses of the geometric approach on a 2 minute walk, (i.e., 15 seconds each). The geometric approach achieved an overall accuracy of 89.87\% for each pose category: standing still (98.15\%), fall forward (85.23\%), fall backward (60.02\%), fall left (91.12\%), fall right (90.22\%), lifting left hand (98.45\%), lifting right hand (98.10\%), and sitting (97.67\%). The highest accuracies were achieved with the standing still position, due to its similarity to the calibration phase, and the lifting hand poses, mainly due to the use of specific wrist markers and the combination with force sensor information.

\subsection{F1 Score for Models Comparison}
\begin{figure}[t]
    \centering
    \includegraphics[scale=0.7]{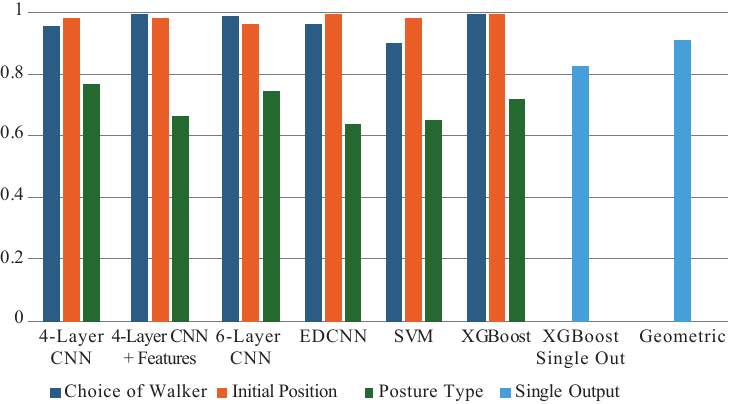}
    \caption{Comparison of F1 Scores for models predicting choice of walker, initial position, and posture type, including single output for XGBoost.}
    \label{fig:f1score}
    \vspace{-0.2cm}
\end{figure}

Fig. \ref{fig:f1score} illustrates the comparison of the F1 Score for the prediction of unseen data. Outputs related to the choice of walker and initial position (e.g., sitting or standing) exhibit the best performance. This is expected due to the number of markers that are visible for these cases. Geometric and XGBoost showed the higher metrics. Moreover, XGBoost with a single output showed 83\% F1 Score. This is a promising result as it does not focus on individual posture classification, but risk factor classification.

\section{Discussion}
In terms of training and validation, XGBoost demonstrated the best overall training and validation performance, achieving 100\% accuracy for the choice of walker task, 99.94\% for initial position, and 99.24\% for posture type, making it the most effective model across both binary and multi-class classification tasks. For the choice of walker and initial position tasks, deep learning models displayed comparable performance, with only slight differences. EDCNN achieved 99.61\% accuracy for choice of walker and 99.25\% for initial position, closely followed by the 4-layer CNN at 99.73\% and 98.97\%, respectively. The 4-layer CNN with extracted features achieved 98.81\% and 97.14\%, while the 6-layer CNN recorded 99.09\% and 97.85\% for these tasks.

However, the posture type classification posed more challenges. The accuracy for this multi-class task was lower across all models, with EDCNN achieving 90.88\%, 4-layer CNN at 90\%, and 4-layer CNN with extracted features at 81.02\%. The 6-layer CNN performed slightly better than the 4-layer CNN with extracted features, with an accuracy of 85.35\%. This suggests that while these models excel in binary tasks, they struggle more with the complexity of multi-class posture classification, particularly with increased architectural complexity like the 6-layer CNN.

Regarding generalization to unseen data, XGBoost is the top performer for binary classification tasks, with nearly perfect accuracy for choice of walker (99.84\%) and initial position (99.69\%). However, the 4-layer CNN delivered the best performance for posture type classification, achieving 76\% accuracy, which was slightly higher than the other models. Notably, the accuracy for posture classification was relatively close across the models, with XGBoost at 74.06\% and other deep learning models ranging between 66\% and 76\%. Although deep learning models performed slightly lower in binary classification tasks, they still demonstrated competitive performance overall.

The geometric posture classification system achieved 98.15\% in standing detection and 89.87\% overall accuracy. The threshold-based detection system performed particularly well for single-user applications, highlighting the potential benefits of personalizing the system for individual users.

As illustrated in Fig. \ref{fig:f1score}, all models achieved F1 scores around 98\% for the choice of walker, as well as for initial position (e.g., standing and sitting). This is a promising result as the implementation of these models could benefit human-robot interaction strategies in smart walkers by understanding when the users are closely in front of the walker or sitting in front of it. In a practical scenario, this information is paramount as it is useful to determine when to trigger the device's brakes in sit-to-stand transfers, for instance. Moreover, for standing detection, this can be contrasted with pressure information from the device's handles, and thus determining if the user is properly using the walker. In terms of posture classification, XGBoost exhibited an 83\% F1 Score and 85\% accuracy for bad posture detection, and the geometric approach achieved 90.8\% F1 Score. In a practical scenario, detecting a posture that differs from standing could potentially indicate a risk factor for falling, i.e., leaning the torso or lifting hands from handles.

Other studies have tackled posture detection using machine-learning techniques. Rivera-Romero \textit{et al.} employed a SVM for posture classification \cite{rivera2024}, but their approach focused on in-bed posture detection with only four postures. Additionally, their use of multiple binary classifiers could hinder real-time performance. Similarly, Stern \textit{et al.} presented a 3D CNN to classify heat map videos of three main in-bed postures \cite{Stern2023}. Although obtaining accuracies around 99\%, this approach requires more computational power and is not strictly related to walker-assisted gait. Oluwaseun \textit{et al.} proposed a solution based on a multi-layer perceptron with transfer learning and hyperparameter optimisation \cite{Ogundokun2022}. Although this method achieved an average accuracy of 89.9\%, it relied on a general-purpose dataset with 410 activities, and would need significant adaptation to work effectively with posture data captured by a walker's camera. Similar deep-learning approaches have been proposed \cite{Wang2021}, but they often rely on general datasets, limiting their applicability to walker-assisted gait scenarios. Pohl \textit{et al.} also presented a multimodal approach to classify gait activities and body posture using videos and motion sensor data \cite{Pohl2022}.

In terms of posture classification based on pressure data, Fonseca \textit{et al.} conducted a literature review of machine-learning and deep learning models, which reported accuracies from 70\% to 99\%, using both wearable and external devices \cite{Fonseca2023}. Regarding other works focusing on posture classification with wearable sensors, Jayasinghe \textit{et al.} implemented a k-Nearest Neighbours algorithm for sitting, standing, lying down, and sitting on the floor, using an IMU with 100\% accuracy \cite{Jayasinghe2023}. However, this work is limited to the classification of data into four basic postures and with wearable devices, limiting is application in older adults.

Several limitations must be acknowledged. The models' performance in multi-class posture classification was notably lower, indicating a need for more sophisticated approaches to handle the complexity of diverse postures. Second, the dataset used was small, limiting the models' generalization. Third, the computational requirements for training deep learning models, such as CNNs, may hinder their practical implementation in resource-constrained environments like single-board computers. Also, the study primarily focused on accuracy metrics, and other aspects like latency and computing power consumption were not assessed.

\section{Conclusions}
This study evaluated various supervised learning models and a traditional approach for classifying postures. The geometric approach achieved the best performance in a reduced set of postures, with 89.9\% accuracy. XGBoost also demonstrated great performance, achieving near-perfect accuracy for binary classification tasks (99.84\% for choice of walker and 99.69\% for initial position) and one of the highest accuracies for posture type classification (74.1\%). Deep learning models, such as the 4-layer CNN and EDCNN, also performed well, particularly in binary tasks, with accuracies above 98\%. However, all supervised learning models faced challenges with the multi-class posture classification task, with lower accuracies across the board. 

Future research should address these limitations by exploring advanced model architectures and feature engineering techniques. Real-world validation in diverse settings is crucial to ensure the models' robustness and applicability. Furthermore, integrating additional data sources, such as pressure sensors and inertial measurement units, could provide a more comprehensive understanding of users' postures and enhance fall detection capabilities. Lastly, evaluating the models' performance concerning computational efficiency and real-time processing will be essential for practical deployment in smart walkers.

\AtNextBibliography{\footnotesize}
\printbibliography

\end{document}